\documentclass[11pt]{article}
\usepackage{galinkin}

\DeclareMathOperator*{\argmin}{\arg\!\min}

\begin{document}

\title{Robustness and Usefulness in AI Explanation Methods}
\author{Erick Galinkin\thanks{erick\_galinkin@rapid7.com}}
\date{\today}

\maketitle

\begin{abstract}
Explainability in machine learning has become incredibly important as machine learning-powered systems become ubiquitous and both regulation and public sentiment begin to demand an understanding of how these systems make decisions.
As a result, a number of explanation methods have begun to receive widespread adoption.
This work summarizes, compares, and contrasts three popular explanation methods: LIME, SmoothGrad, and SHAP.
We evaluate these methods with respect to: robustness, in the sense of sample complexity and stability; understandability, in the sense that provided explanations are consistent with user expectations; and usability, in the sense that the explanations allow for the model to be modified based on the output.
This work concludes that current explanation methods are insufficient; that putting faith in and adopting these methods may actually be worse than simply not using them.
\end{abstract}

\section{Introduction} \label{sec:intro}
Machine learning has become a ubiquitous part of life for us.
These algorithms aim to predict, either via classification or regression, information about a particular input.
The trouble comes in that, as Doshi-Velez and Kim~\cite{doshi2017towards} put it, any single metric is necessarily an incomplete description of the real world.
Given the influence these systems have in our lives, people seek to understand precisely why a model made a particular decision.
To this end, research efforts around interpreting and explaining the predictions of models have sprung up and have fallen into two overarching families: intrinsically interpretable models and post-hoc explanation methods. 
The results of the methods are typically quite similar, but while interpretable models aim to be sufficiently simple that their predictions are easily reasoned about, post-hoc explanation methods take a trained model that may be extremely complex and construct explanations from properties of the model.
This work focuses primarily on popular explanation methods applied to trained models rather than inherent interpretability.

Explanation methods are of particular concern because of the way that they influence human behavior.
Despite the challenges in defining terms like interpretability in concrete terms~\cite{lipton2018mythos}, explanations are important for improving trust and can also improve the speed at which human auditors perform their work~\cite{schmidt2019quantifying}.
For some users, the mere presence of an explanation increases that user's trust in a model, even when the explanations are incorrect~\cite{bansal2021does}.
In other cases, even data scientists who build models tend to rely too heavily on these interpretability tools~\cite{kaur2020interpreting}, demonstrating a failure to use the tool's appropriately.
These challenges, and others, are detailed in Section~\ref{sec:critique}.

\section{Background}
This work reflects on a number of explanation methods -- namely LIME, SmoothGrad, and SHAP.  
These methods have begun to see adoption in industry~\cite{gade2019explainable} and are popular for their ease of implementation.
To facilitate the discussion in Section~\ref{sec:comparison}, we introduce the motivation and implementation of these methods below.

\subsection{Local Interpretable Model-Agnostic Explanations (LIME)}
Local Interpretable Model-Agnostic Explanations (LIME)~\cite{ribeiro2016should} is an explanation method that aims to explain the predictions of any classifier, irrespective of complexity or linearity.
This is accomplished through an interpretable representation that locally approximates the overall classifier.
The interpretable representation is derived from a class of potentially interpretable models -- limited in part by the complexity of the model, denoted $\Omega(g)$.
In this case, examples of complexity include tree depth or the number of non-zero weights of a linear model.

In addition to minimizing complexity, LIME also seeks a model that minimizes the locality-aware loss, or so-called fidelity function $\mathcal{L}$ for an input $x$. 
$\mathcal{L}$ is a function that measures how good an interpretable model $g$ is at approximating a classifier $f$ in a neighborhood $\pi_x$ of $x$.
Formally, given an input $x$, a classifier $f$, and a class of potentially interpretable models $G$, the explanation produced by LIME is obtained by:
\begin{equation} \label{eqn:LIME}
\xi(x) = \argmin_{g \in G} \mathcal{L}(f, g, \pi_x) + \Omega(g)	
\end{equation}

LIME then perturbs and samples the space around each instance of interest, $x$, constructing representations and labels for those representations.
The new samples are weighted by their distance from $x$.
These representations and labels are then used to select and train the model satisfying Equation~\ref{eqn:LIME}.
This means that the explanation model used for a particular instance $x$ is the model $g$ that is closest to the classifier $f$ with minimal model complexity $\Omega(g)$.
In practice, $\Omega(g)$ is a hyperparameter~\cite{molnar2020interpretable}.
For example, if we want to use linear regression models as our locally interpretable model, we select the number of parameters $K$ to be used in the model.

\subsection{SmoothGrad} \label{sec:smoothgrad}
SmoothGrad~\cite{smilkov2017smoothgrad} is a gradient-based explanation method that builds on work like Saliency Maps~\cite{simonyan2013deep} and GradCAM~\cite{selvaraju2017grad}.
Crucially, SmoothGrad is not, on its own, an explanation method, but rather an extension of any existing gradient-based explanation method.
When we refer to SmoothGrad on its own, we refer to implementations like iNNvestigate~\cite{alber2018iNNvestigate} where it complements so-called ``vanilla'' gradient methods.

These gradient-based methods take advantage of the backpropogation algorithm used in training neural networks, particularly convolutional neural networks used in image classification. 
SmoothGrad was developed in response to two issues in gradient-based explanation methods:
\begin{enumerate}
	\item The issue of one feature with a strong global effect but a small local derivative ``saturating'' the sensitivity map
	\item Noisy gradients as a result of the class sensitivity function's derivative fluctuating locally
\end{enumerate}

SmoothGrad aims to solve these problems by generating multiple versions of an image of interest $x$ by adding noise to it.
Pixel attribution maps -- heatmaps like those generated in Selvaraju \textit{et al.}~\cite{selvaraju2017grad} -- are then created for all of these images, and those maps are then averaged together.
The idea here is that adding some noise sampled from a Gaussian distribution averaged over multiple maps can smooth the local gradient fluctuations and lead to better maps. 

\subsection{Shapley Additive Explanations (SHAP)}
Shapley Additive Explanations (SHAP)~\cite{lundberg2017unified} is an explainability method that leverages a game theoretic approach to understand model predictions.
The work was motivated as an attempt to unify six different methods for interpreting model predictions, including LIME.
SHAP works by assigning an importance value to each feature for each prediction using Shapley value estimation methods.
The Shapley value~\cite{shapley1953quota} is a construct from cooperative named for Lloyd Shapley that seeks to distribute the total gain in utility across a coalition of $n$ players according to each player's marginal contribution.
Formally, given a characteristic function $v: 2^n \rightarrow \mathbb{R}, v(\emptyset) = 0$ that maps a a set of players to a value for that coalition, the set of all players $N$ and a coalition of players $S \subseteq N$, a fair distribution for each player $i$ in a coalition $S$ is given by:
\begin{equation} \label{eqn:SHAP}
\phi_i(v) \sum_{S \subseteq N \setminus \{i\}} \frac{\left|S\right|! (\left|N\right| - \left|S\right| - 1)!}{\left|N\right|!} \left(v(S \cup \{i\}) - v(S)\right) 	
\end{equation}

Theorem 1 of Lundberg and Lee~\cite{lundberg2017unified} shows that only one possible explanation model for the properties of local accuracy, missingness, and consistency holds for additive feature attribution methods -- explanation models that are linear functions of binary variables -- the Shapley value.
Despite the fact that exactly computing Shapley values is NP-Hard, the paper develops model-agnostic approximations by assuming feature independence and model linearity.
The paper describes two classes of approximations: model-agnostic approximations and model-specific approximations. 
The model-agnostic approximation introduced in this initial work is called Kernel SHAP, which leverages linear LIME with Shapley values.
The model-specific approximations in the initial paper handle linear models and deep neural networks using techniques from DeepLIFT~\cite{shrikumar2017learning}.

The initial model was extended to provide visual explanations in later work~\cite{lundberg2018explainable}. 
A further extension of SHAP~\cite{lundberg2020local} tackled the problems associated with model-specific approximations of nonlinear, tree-based models.
Tree SHAP, as the author calls it, leverages feature perturbations to locally estimate the interaction index, a game theoretic concept that follows from generalization of the Shapley value properties.

\section{Comparison of Explainability Methods} \label{sec:comparison}
The three explainability methods above are all post-hoc methods that give a view into machine learning black boxes.
As discussed in Section~\ref{sec:intro}, there is growing interest in leveraging explainability and interpretability methods in practice. 
This means that is is prudent to understand the strengths and shortcomings of each method, and leverage the right one for our task.
Table~\ref{tab:summary} summarizes some of the commonalities and differences between the methods, which we detail in this section.

\begin{table}[h]
\begin{tabular}{l|ccccc}
           & Methodology  		& Local Explanation 		& Visualizable 	& Model Agnostic  \\\hline
LIME       & Perturbation       & \checkmark  			& X 				& \checkmark 	 \\
SmoothGrad & Gradient           & \checkmark 			& \checkmark 	& X 	 		 \\
SHAP       & Perturbation       & \checkmark 			& \checkmark  	& \checkmark	
\end{tabular}
\caption{Summary of LIME, SmoothGrad, and SHAP}
\label{tab:summary}
\end{table}

\subsection{Methodology}
The three methods we consider fall under two different methodologies -- perturbation and gradient.
Perturbation methods add some noise to inputs and sample the space of perturbed inputs to assess the effect of small changes in the space.
Gradient-based methods instead use the gradients computed for given inputs to explain model predictions.

Agarwal \textit{et al.}~\cite{agarwal2021towards} consider a variant of LIME for continuous data called C-LIME and compare it directly to SmoothGrad. 
Interestingly, though ado much has been made of the differences between perturbation and gradient-based explanations, Agarwal \textit{et al.} demonstrate that given the same Gaussian distribution for perturbations and gradient computation, C-LIME and SmoothGrad yield the same explanation in expectation.
This suggests that the methodology of the explanation is, on average, not meaningful to distinguish methods.

\subsection{Local Explanation}
We define local explainability as the ability to interrogate and explain particular examples from the data. 
All three of the methods we consider here offer local explanations for single predictions.
This is a property that is often desirable for legal and regulatory reasons, since it is often individual predictions that lead to negative outcomes for individuals.

\subsection{Visualizable}
Visualizations are valuable tools for communicating data concisely in ways that humans quickly and easily understand~\cite{tufte1998visual}.
Consequently, having explanations methods that lend themselves to visualizations can improve understanding and interpretation.
SmoothGrad can be used with any number of gradient-based methods, but its use with the so-called ``vanilla'' gradient as mentioned in Section~\ref{sec:smoothgrad} lends itself to the creation of heat maps. 

SHAP, though it does not produce heat maps on images, has a Python package~\footnote{https://github.com/slundberg/shap} that generates force plots which show how each feature contributes to the model output.
This method is used for individual predictions but can also be used across entire datasets to understand how the model treats the entirety of the dataset.
In cases where we want to understand the influence of a single feature, SHAP offers a way to visualize the impact of that single feature across the dataset.

LIME, on the other hand, does not lend itself to a natural visualization. 
The relatively simple nature of the LIME model means that variable importance measures or partial dependence plots can be used to visualize attributes of the more interpretable model, but these visualizations are less direct than those produced by SmoothGrad or SHAP.
Specifically, the visualization would be informative about the surrogate model around the point of interest, and would not give a view into how the larger model made the prediction relative to the input in question.

\subsection{Model Agnostic}
Model agnostic methods are able to be used across a variety of model types. 
By the No Free Lunch Theorem~\cite{wolpert1997no}, we know that there is no single model that is optimal for all tasks. 
This means that in practice, different models will be used for different tasks and those models may require explanation in different ways.
Moreover, the way that data is input to the model can vary wildly -- images and video data structures often look quite different from text, and all of these require different handling and processing methods from tabular data.
Model agnostic explanation methods aim to handle any of these input methods.

LIME specifically aims to be model-agnostic and achieves this by uncovering locally-interpretable models. 
SHAP, building on the foundation of LIME, also inherits this property.
Though some specifics between \textit{e.g} KernelSHAP~\cite{lundberg2017unified} and TreeExplainer~\cite{lundberg2020local} differ, the core of the methods remain the same. 
This means that the skills associated with interpreting \textit{e.g.} Shapley values are transferrable across models.

SmoothGrad is not model agnostic and is only usable within differentiable models.
In fact, SmoothGrad is specifically useful in the case of convolutional neural networks used for computer vision applications.
This makes it a powerful tool for identifying where in an image the model is ``paying attention'', but limits the usefulness of the method in general.

\section{Weaknesses of Explainability Methods} \label{sec:critique}
Model trust is a critically important concept, gaining the attention of many, including the National Institute of Standards and Technology~\cite{stanton2021trust} and the parliament of the European Union~\cite{eu2021proposal}.
This concept of trust, however, builds on a shaky foundation.
In particular, nearly all legislation proposed to date includes a requirement for model interpretability. 
Lipton~\cite{lipton2018mythos} offers a stunning rebuke of the term itself, as the term has no agreed upon meaning.
In essence, the question becomes not whether a model is interpretable or not, but rather to whom and under what circumstances model predictions can be understood.

\subsection{Human Factors in Explanations}
One critique leveled at fairness in machine learning~\cite{fazelpour2020algorithmic} is the notion that one must conceive of an ``ideal world'' against which the current world is compared. 
This can lead to unexpected negative outcomes for certain groups and notions of fairness that leave everyone worse off: for example, a perfectly fair judge who always denies bail and parole.
The weaknesses of metrizable fairness then promotes the desire for explanations -- \emph{why} was bail denied?

A bevy of work has shown that although explainability methods have been widely promoted, the explanations provided by the likes of LIME, SmoothGrad, and SHAP can also range from being useless to outright harmful.
Kaur \textit{et al.}~\cite{kaur2020interpreting} show that data scientists tend to misuse and over-rely on explanations from tools like SHAP. 
Moreover, decision-makers often defer to algorithmic decision support systems~\cite{green2019principles} and struggle to use the algorithms effectively -- often underperforming compared to both humans who are not assisted and the algorithms themselves~\cite{lai2019human}.  
Model-agnostic methods are not alone here, as Grad-CAM~\cite{selvaraju2017grad} has been shown to perform quite poorly on tasks it was specifically designed to excel at~\cite{prakash2022towards} under benchmark conditions that are not sufficiently realistic.

Explainability often uses healthcare for piloting its methods, likely due to the superhuman performance of machine learning methods on a variety of medical tasks~\cite{choudhury2020role,oren2020artificial,hosny2018artificial}.
SHAP in particular has been used in healthcare contexts~\cite{lundberg2018explainable} and has been touted as a machine learning explainability method with promise to improve healthcare worker trust in algorithms.
This is very likely to be true, as even when users are presented explanations are wrong~\cite{bansal2021does} or completely random~\cite{springer2017dice}, they tend to believe that the algorithm is correct.
Indeed, in the healthcare imaging context~\cite{ghassemi2021false}, localizing a region of an image does not inform a human exactly \emph{what} in that region the model considered useful.
Combining these two phenomena -- a lack of clarity in the explanation paired with human deference to decision support algorithms -- is ultimately putting a human in the loop purely for decoration in the best case.

\subsection{Robustness of Explainers}
Beyond the human factors, the theoretical properties of explanation methods leaves much to be desired.
Agarwal \textit{et al.}~\cite{agarwal2021towards} investigate robustness for C-LIME and SmoothGrad in certain settings, but find that the robustness is sensitive to variance in the data.
Robustness is a particularly valuable attribute for explanations, as humans expect the important factors for data that are ``near'' each other to be similar. 
Despite the equivalence of perturbation and gradient-based methods in expectation, perturbation-based methods are empirically worse~\cite{alvarez2018robustness} at providing consistent explanations given small modifications to the input.
Moreover, since there are so many models and explanation methods that achieve comparable performance, this phenomenon may lead to selecting models on the basis of how pleasing the explanations are~\cite{hancox2020robustness} without regard for the consistency of those explanations.
Rahnama and Bostr\"{o}m~\cite{rahnama2019study} also find that LIME is not robust to changes in the distribution of the data -- a common occurrence in production applications of machine learning.

This lack of robustness also allows for explainers to be ``tricked''~\cite{slack2020fooling} by models that aim to offer pleasing explanations, even when those explanations do not provide visibility into the model itself.
In image classification, similarly, so-called ``adversarial examples''~\cite{szegedy2013intriguing} -- inputs with specially-crafted perturbations that are imperceptible to humans -- can be crafted to arbitrarily change saliency maps~\cite{dombrowski2019explanations,heo2019fooling}.
As a result of this, a model designer who is for whatever reason disinterested in removing bias from their model is able to provide explanations with whatever properties they wish -- a property that  Aivodji~\textit{et al.}~\cite{aivodji2019fairwashing} have termed ``fairwashing''.

Model explanations are, of course, also impacted by human behavior that is used as input.
In recent work, Arora \textit{et al.}~\cite{arora2022explain} ask participants to edit hotel reviews in such a way that a human reader can still derive whether a review is still positive, negative, or mixed, while model confidence for the predicted class is minimized.
The study finds that for BERT~\cite{devlin2018bert}, a transformer model widely used in natural language processing, local explanations via LIME and Integrated Gradients offer no additional ability for participants to reduce the confidence of the model, while a linear model trained only on the predictions of the BERT model offered participants significantly more ability to change the label associated with the sample.
Ultimately, this demonstrates that the relative utility of the local explanations is lower than simply using feature coefficients from a linear model that approximates the global model.

\section{Conclusion}
Post-hoc explanation methods offer significant opportunity to improve the usability and human comprehensibility of machine learning methods.
However, current popular methods face a variety of problems from both a theoretical and sociotechnical perspective. 
Though these problems may not be unsolvable or inherent, further investigation is needed on how to increase the robustness and usefulness of explanations.
Given the concomitant phenomena of human deference to decision support systems, human willingness to accept incorrect explanations, a lack of robustness for explanations, and a lack of theoretical guarantees for extant explanation methods, the idea of a ``human-in-the-loop'' serves as nothing more than a security blanket.
In order to deal with model interpretability and explainability, more foundational work needs to be done to formalize problems in interpretability and consider ideas around causality and contestability. 
Counterfactual explanations offer a potential path forward, though extant counterfactual generation methods have their own weaknesses and biases.

\bibliographystyle{acm}
\bibliography{references}

\end{document}